\documentclass[10pt,twocolumn,letterpaper]{article}

\usepackage{xr}
\makeatletter
\newcommand*{\addFileDependency}[1]{
  \typeout{(#1)}
  \@addtofilelist{#1}
  \IfFileExists{#1}{}{\typeout{No file #1.}}
}
\makeatother

\newcommand*{\myexternaldocument}[1]{
    \externaldocument{#1}
    \addFileDependency{#1.tex}
    \addFileDependency{#1.aux}
}

\usepackage{wacv}
\usepackage{times}
\usepackage{epsfig}
\usepackage{graphicx}
\usepackage{amsmath}
\usepackage{amssymb}
\usepackage{booktabs}

\usepackage{url}
\usepackage{nicefrac}
\usepackage{microtype}      
\usepackage{xcolor}         
\usepackage{multicol}
\usepackage{lipsum}
\usepackage{mwe}
\usepackage[utf8]{inputenc}
\usepackage[english]{babel}
\usepackage{easyeqn}
\usepackage{bbm}
\usepackage[thinlines]{easytable}
\usepackage{caption}
\usepackage{subcaption}
\usepackage{multirow}
\usepackage{dsfont}
\usepackage{textcomp}
\usepackage[T1]{fontenc}

\usepackage{algorithmicx}
\usepackage{algorithm}
\usepackage{algpseudocode}
\usepackage[algo2e]{algorithm2e}

\myexternaldocument{supplementary_material}

\usepackage{xr-hyper}

\def\wacvPaperID{2048} 

\wacvalgorithmstrack   

\wacvfinalcopy 

\ifwacvfinal
\usepackage[breaklinks=true,bookmarks=false]{hyperref}
\else
\usepackage[pagebackref=true,breaklinks=true,colorlinks,bookmarks=false]{hyperref}
\fi

\def\t{\times}

\def\RR{\mathbb{R}}

\def\ev{\mathbf{e}}

\def\uv{\mathbf{u}}
\def\wv{\mathbf{w}}
\def\xv{\mathbf{x}}
\def\yv{\mathbf{y}}

\def\epsv{\boldsymbol{\varepsilon}}

\def\lv{\mathbf{l}}

\newcommand{\sS}{N} 
\newcommand{\cN}{C} 
\newcommand{\sigm}{\mathrm{sigm}}

\begin{document}

\title{ScaleFace: Uncertainty-aware Deep Metric Learning}

\author{
Roman Kail\thanks{These authors contributed to research equally}\\
Skoltech, Sberbank\\
Moscow, Russia\\
{\tt\small roma.vkail@gmail.com}
\and
Kirill Fedyanin$^*$\\
Technology Innovation Institute\\
Abu Dhabi, United Arab Emirates\\
{\tt\small kirill.fedianin@tii.ae}
\and
Nikita Muravev\\
Lomonosov Moscow State University\\
{\tt\small ne-ki-tos@yandex.ru}
\and
Alexey Zaytsev\\
Skoltech\\
Moscow, Russia\\
{\tt\small a.zaytsev@skoltech.ru}
\and
Maxim Panov\\
Technology Innovation Institute\\
Abu Dhabi, United Arab Emirates\\
{\tt\small maxim.panov@tii.ae}
}

\maketitle
\thispagestyle{empty}

\begin{abstract}
  The performance of modern deep learning-based systems dramatically depends on the quality of input objects. For example, face recognition quality would be lower for blurry or corrupted inputs. However, it is hard to predict the influence of input quality on the resulting accuracy in more complex scenarios. We propose an approach for deep metric learning that allows direct estimation of the uncertainty with almost no additional computational cost. The developed \textit{ScaleFace} algorithm uses trainable scale values that modify similarities in the space of embeddings. These input-dependent scale values represent a measure of confidence in the recognition result, thus allowing uncertainty estimation. We provide comprehensive experiments on face recognition tasks that show the superior performance of ScaleFace compared to other uncertainty-aware face recognition approaches. We also extend the results to the task of text-to-image retrieval showing that the proposed approach beats the competitors with significant margin.

\end{abstract}

\section{Introduction}
\label{sec:intro}
    
  Deep metric learning~\cite{kaya2019deep,hav4ik2021deepmetriclearning} is currently the leading approach to perform machine learning in such challenging scenarios as open-set classification~\cite{gunther2017toward,masone2021survey} and object retrieval~\cite{radford2021learning}. Unlike the standard closed-set classification, the above-mentioned problems require the models to work with classes different from the ones used during training as the number of classes is huge, and new ones emerge every day.


  The standard approach in deep metric learning is to use so-called \emph{backbone} model that produces embeddings. Then one compares the obtained embeddings to decide if a corresponding pair of objects belong to one class or not. More formally, it performs one-nearest-neighbor classification for some distance between embeddings. 

  In this work, we focus on uncertainty estimation for open-set recognition models. Uncertainty estimation methods target assessing the confidence in the prediction for particular input objects. 
  We argue that uncertainty estimates for such models are of high importance. 
  For example, face recognition systems can report high similarity score not only for images with the same identity, but for low-quality (blurry, noisy, ...) images of different identities thus producing a false positive result, see~\cite{Shi2019ProbabilisticFE}. 

  Importantly, the majority of existing uncertainty estimators are designed to work in a more traditional closed-set classification scenario. In this task, training and test sets of objects share the same set of classes. For closed-set uncertainty estimation, one can use output probabilities as a strong baseline~\cite{mukhoti2021deterministic}. 
  %
  This paper shows that existing open-set pipelines, where instead of probabilities of classes, we have distances between objects, make the problem of uncertainty estimation more tricky. 
  The quality benefits provided by existing approaches~\cite{Shi2019ProbabilisticFE,meng2021magface,chen2022fast} are usually moderate while computational complexity is often much higher than standard methods like ArcFace~\cite{deng2019arcface} or CosFace~\cite{wang2018cosface}.
  
  In this work, we develop a new method for deep metric learning \textit{ScaleFace} that aims to provide computationally efficient uncertainty estimates simultaneously improving the downstream task quality. The idea is to make a scale value in ArcFace loss function~\cite{deng2019arcface} to be an object-specific quantity. The approach requires only a small modification of existing metric learning pipelines as scale value can be computed by a small separate head of a network.
  Thus, both training and inference time for ScaleFace is almost identical to the vanilla ArcFace. 
  %
  The key contributions of this work are as follows.
  \begin{enumerate}
    \item We introduce a new deep metric learning method \textit{ScaleFace} that has natural uncertainty estimation capabilities and is computationally efficient, see Section~\ref{sec:open_set_methods} for the description of the method. 
    
    \item We perform a careful experimental evaluation of ScaleFace on face recognition problems in Section~\ref{sec:open_set_experiments}, and show that it outperforms the competitors.

    \item We extend the method to the problem of text-to-image retrieval and show its efficiency in this task, see Section~\ref{sec:retrieval}.
  \end{enumerate}
  Additional details and results are available in Supplementary Material (SM).

\section{Methods}
\label{sec:open_set_methods}

\begin{figure*}[t!]
    \includegraphics[width=0.8\textwidth]{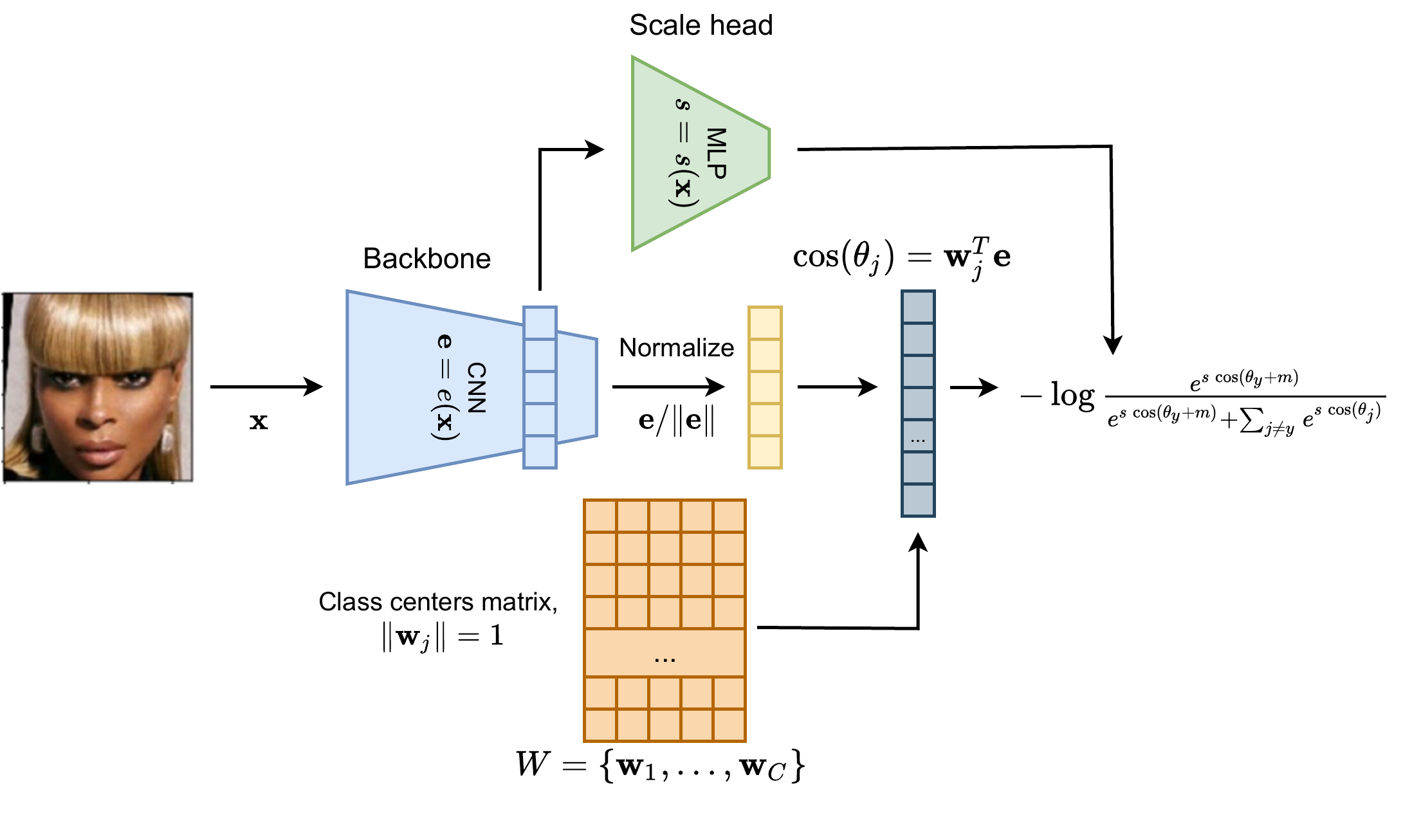}
    \centering
    \caption{Pipeline of training scale to predict uncertainty. We have a two-headed backbone, that predicts the embedding vector $\ev(\xv_i)$ and the scale coefficient $s(\xv_i)$. Then these two values are processed the same way as in ArcFace article to get the loss function. The obtained scale head allows fast uncertainty estimation during inference.}
  \label{fig:scale_prediction_pipeline}
\end{figure*}

\subsection{ArcFace model}
\label{sec:arcface_intro}
  In a training dataset $\{(\xv_i, y_i)\}_{i = 1}^{\sS}$, $\xv_i$ is a description of an object (for example, RGB image) and $y_i \in \{1, \dots, \cN\}$ is class label. Here $\cN$ is the total number of classes in the training data. 
  We consider models that transform $\xv$ into an embedding $\ev(\xv) \in \RR^{d}$ via an encoder network. Here $d$ is an embedding dimension (usually, equal to $512$).

  The standard classification loss function is a softmax loss, given by the following formula:
  \begin{equation}
    L = - \frac{1}{N} \sum_{i = 1}^{N} \log \frac{e^{\langle \ev_i, \wv_{y_i}\rangle}}{\sum_{j = 1}^{\cN} e^{\langle \ev_j, \wv_{j}\rangle}},
  \end{equation}
  where $\wv_j \in \RR^d$ is the centroid vector for the class $j$, $\ev_i = \ev(\xv_i) \in \RR^d$ is an embedding of the $i$-th object by an encoder. We absorb the bias term in vectors $\wv_j$ to simplify the notation. Essentially, if we denote $W = \{\wv_j\}_{j = 1}^C$ then the logits of classes for object $\xv_i$ are given by $\lv_i = W \ev_i$ with $W \in \RR^{C \t d}$ being parameters of the last (fully connected) layer of a network.

  The ArcFace model~\cite{deng2019arcface} suggests to normalize both embedding vectors $\ev_i$ and class centroids $\wv_j$ to have unit l2-norm. As a result, scalar product $l_{ij} = \langle  \ev_i, \wv_{j} \rangle$ boils down to the cosine similarity between embedding and class vectors: $l_{ij} = \cos \theta_{ij}$, where $\theta_{ij}$ is the angle between vectors $\ev_i$ and $\wv_{j}$. 
  The vector of logits is multiplied by \emph{scale} constant~$s$.
  In the original article, the scale equals $64$. 
  The resulting vector is passed to the softmax function and then to the cross-entropy loss function:
  \begin{equation}
    L = - \frac{1}{N} \sum_{i = 1}^{N} \log \frac{e^{s \cos (\theta_{i y_i } + m)}}{e^{s \cos (\theta_{i y_i} + m)} + \sum_{j \ne y_i} e^{s \cos (\theta_{ij})}}.
  \label{eq:arcface_loss}
  \end{equation}
  ArcFace also adds margin $m$ to the terms in the loss related to the true class to move classes further away from each other.

  Let us note that if the distribution of probabilities over classes is close to one-hot, that means that the model is confident about its prediction. 
  Otherwise, if classes have almost equal probabilities, the model is uncertain about its decision. We will build on this intuition to propose a modification to ArcFace model in the next section.

\subsection{Prediction of uncertainty using scale}
\label{sec:scale_theory}
  The entropy of the probability distribution of the classes is a strong indicator of prediction uncertainty. In ArcFace pipeline, the scale is the parameter, responsible for the entropy of the resulting distribution. Thus, by adjusting scales for individual examples we can account for uncertainty in a meaningful way.

  To compute object-dependent scale values, we suggest to train an extra head of the network. In our implementation, this subnetwork takes activations from the penultimate layer of the backbone, transforms them via multilayer perceptron, and then predicts the scale coefficient for each image individually. The training pipeline for the network with the scale predicting subnetwork is shown on Figure~\ref{fig:scale_prediction_pipeline}. The training procedure remains the same, only the scale becomes not a hyperparameter, but a value predicted by a separate head of the network. 

  Now consider an input object $\xv_i$ that we process with a two-headed backbone to get the l2-normalized embedding vector $\ev(\xv_i)$ and scale coefficient $s(\xv_i)$. Then we can compute the corresponding modification of~\eqref{eq:arcface_loss} that takes into account object-dependent scale coefficients:
  \begin{EQA}[c]
    L = - \frac{1}{N} \sum_{i = 1}^{N} \log \frac{e^{s(\xv_i) \cos (\theta_{i y_i} + m)}}{e^{s(\xv_i) \cos (\theta_{i y_i} + m)} + \sum_{j \ne y_i} e^{s(\xv_i) \cos (\theta_{ij})}}.
  \label{eq:scale_loss}
  \end{EQA}
  In this setting, the prediction of high values of $s(\xv_i)$ moves the probabilities distribution after softmax closer to the one-hot distribution with value $1$ for the logit with highest value. On the other hand, low values of $s(\xv_i)$ move the probability distribution towards the uniform one. Consider the object for which the model has selected the right class. Then, in order to further minimize the loss, it is beneficial for the model to predict the high value of the scale coefficient. Otherwise, if the model misses the target class, it is beneficial to predict low scale value. This intuition makes $s(\xv_i)$ a reasonable measure of confidence of a network in the prediction. We call the resulting model \textit{ScaleFace}.
  

\subsection{From scaling coefficient to modified similarity measure}
\label{sec:modified_similarity}

  The PFE approach~\cite{Shi2019ProbabilisticFE} suggests not only to compute uncertainties for the input objects but also proposes the modified similarity measure that takes uncertainties into account. This similarity measure, \textit{mutual likelihood score (MLS)}, allows to improve the final quality of the model. In our work, we suggest to use predicted scale coefficients in a modified similarity metric improving over cosine similarity.

  In training procedure that was described in Section~\ref{sec:scale_theory} we were predicting logits, each of them being equal $l_j(\xv) = s(\xv) \langle \ev(\xv), \wv_j \rangle$, where $\ev(\xv)$ is the l2-normalized vector, predicted by the backbone, $\wv_j$ is the centroid vector corresponding to the $j$-th class and $s(\xv)$ is the predicted scale coefficient. This function represents the usual cosine similarity between vectors $\ev(\xv)$ and $\wv_j$ but adjusted by a scale coefficient $s(\xv)$.

  The similarity measure efficiently used during training inspires to modify the one used at inference stage. Here we consider two possible scenarios:
  \begin{enumerate}
    \item We aim to compare two objects $\xv_1$ and $\xv_2$ taking into account uncertainties for both of them. Here we assume that the similarity is used to solve binary classification problem, distinguishing pairs, belonging to one identity (positive class) or different identities (negative class).Then, we suggest to consider the similarity measure 
    \begin{equation}
      s(\xv_1, \xv_2) \langle \ev(\xv_1), \ev(\xv_2) \rangle,
    \label{eq:mod_similarity_pair}
    \end{equation}
    where $s(\xv_1, \xv_2)$ is some function computed based on the scales $s(\xv_2)$ and $s(\xv_2)$. For example, one may consider $s(\xv_1, \xv_2) = \sqrt{s(\xv_1) \cdot s(\xv_2)}$.

    \item We compare query object $\xv$ with some template class object representation $\uv_i$ for which we are sure in the quality of representation. In this case, we simply consider
    \begin{equation}
      s(\xv) \langle \ev(\xv), \uv_i \rangle.
    \label{eq:mod_similarity_template}
    \end{equation}
    Such a similarity measure most closely resembles the one used during training.
  \end{enumerate}
    
  Let us note that while modified similarities~\eqref{eq:mod_similarity_pair} and~\eqref{eq:mod_similarity_template} represent the essential idea of pushing uncertain objects to have low similarity, in practice the decision boundary between classes has substantial positive value. That is why the direct application of these formulas may lead to many positive examples receiving similarities lower than class-separation threshold. In order to overcome this issue, we propose to introduce the shift parameter $\mu > 0$ and consider a modification of similarity measure~\eqref{eq:mod_similarity_pair} of the following form:
  \begin{equation}
    s(\xv_{1}, \xv_{2}) \bigl(\langle \ev(\xv_{1}), \ev(\xv_{2})\rangle - \mu\bigr).
  \label{eq:mod_similarity_eq}
  \end{equation}
  By tuning parameter $\mu$ we can try to push uncertain pairs closer to class separation border and confident pairs further from it. We can modify the template-based similarity measure~\eqref{eq:mod_similarity_template} in the same way. The selection of parameter $\mu$ can be done based on training or validation data (if available),
   see details in SM.

  In Section~\ref{sec:metric_improvement} we will show how the proposed metrics improve the resulting quality of recognition.

\section{Open-Set Experiments}
\label{sec:open_set_experiments}

  \begin{figure*}[t!]
    \centering
    \includegraphics[width=0.8\textwidth]{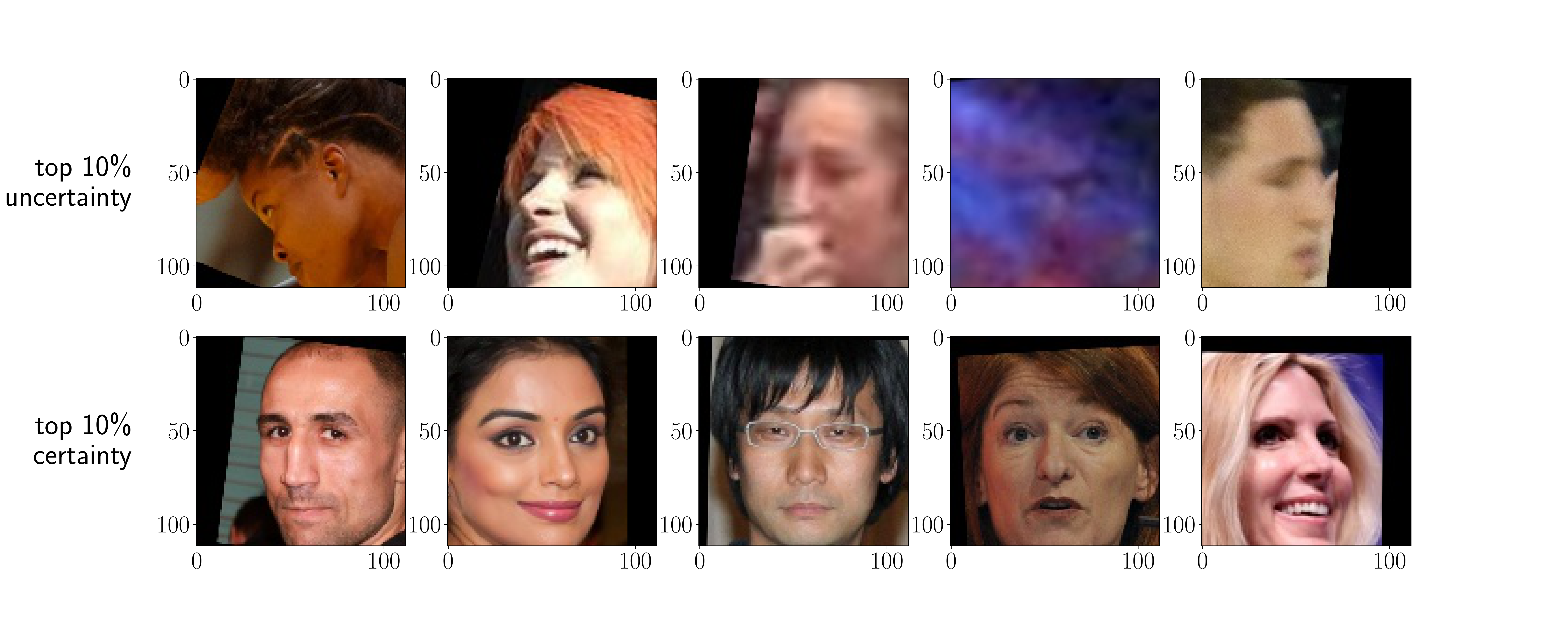}
    \caption{Examples of faces from top $10\%$ and bottom $10\%$ deciles as computed by the introduced scale-based uncertainty.}
  \label{fig:face_examples}
  \end{figure*}

  \begin{figure}[t!]
    \includegraphics[width=0.9\linewidth]{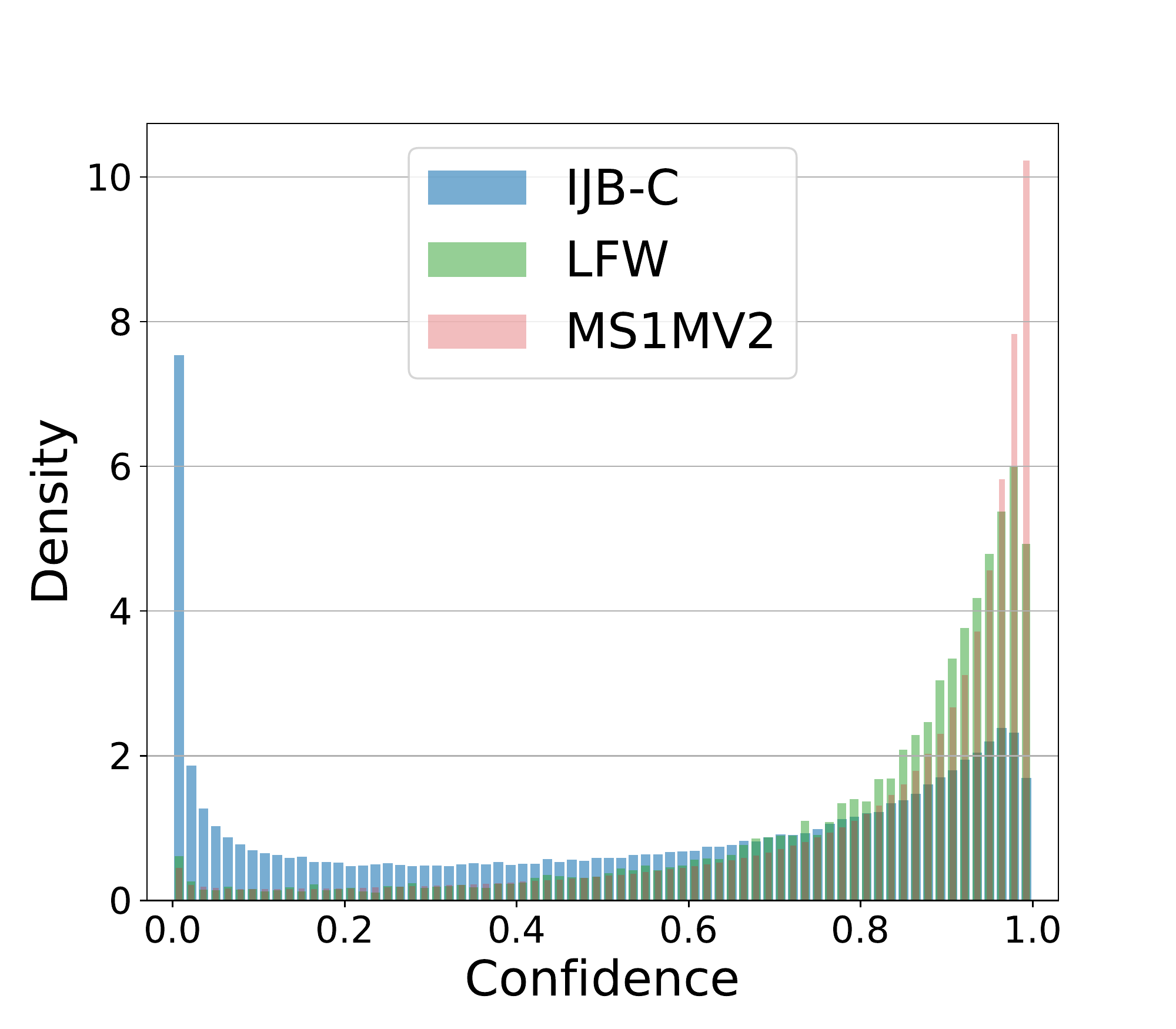}
    \centering
    \caption{Distribution of confidence values produced by ScaleFace. We compare values for three datasets: two validation datasets IJB-C, LFW, and training dataset MS1MV2. To highlight differences, we apply monotonic Box-Cox transformation to initial values.}
  \label{fig:uncertainty_hist_open_set}
  \end{figure}

  \begin{figure*}[t!]
    \includegraphics[width=\textwidth]{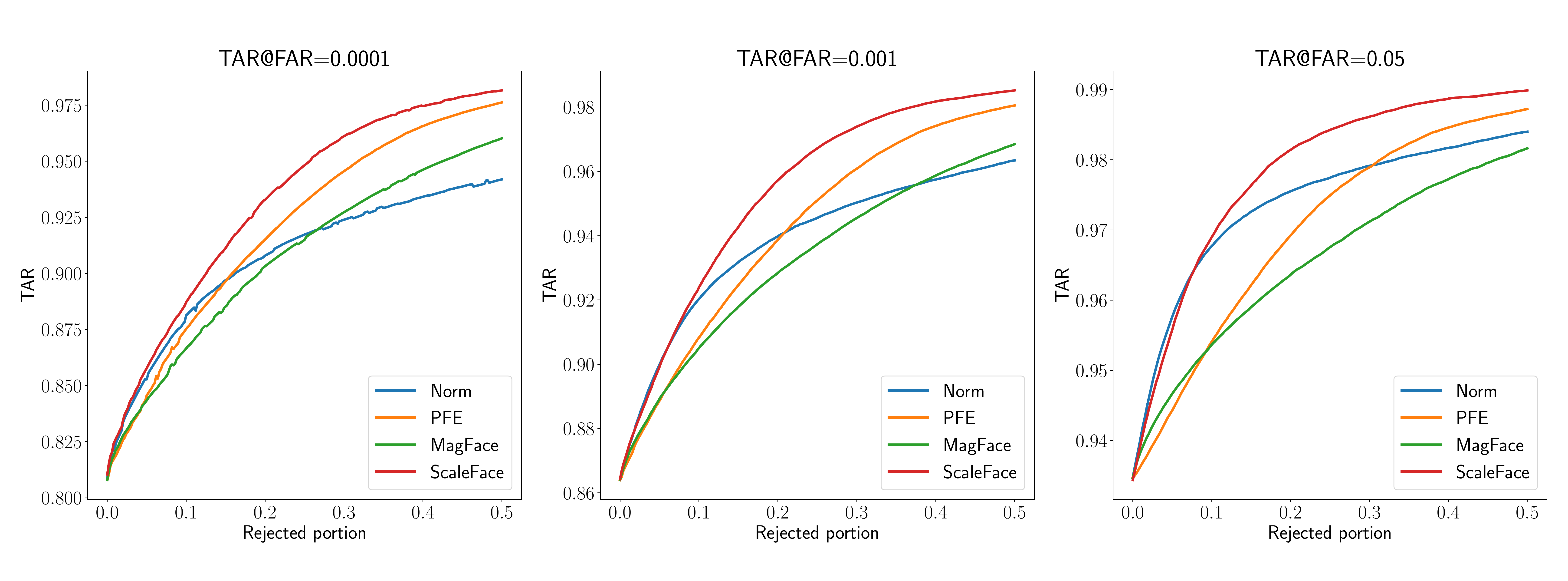}
    \centering
    \caption{Rejection curves for uncertainty estimation in verification task on IJB-C test set. For all of these experiments we use the same ArcFace backbone. Thus we compare the uncertainty estimation performance without regard to the quality of the backbone, that is why curves start in one point.}
  \label{fig:scale_search_activation_2}
  \end{figure*}

\subsection{Overview}
  We start experiments in Section~\ref{sec:qualitative_experiments} with a couple of basic experiments to show that images with higher scale values are easier to recognize with a human eye. Additionally, we show that more complex datasets get lower scales from a model on average.

  Another useful property of uncertainty estimates is that it allows a model to say ``I don't know''. One way to quantitatively estimate it is to drop part of the worst predictions; the faster the key metric grows in this case, the better. As a key metric for face verification, we take commonly used TAR@FAR (true acceptance rate at fixed false acceptance rate) and show the efficiency of the proposed approaches in Section~\ref{sec:reject_verification}.

  Finally, in Section~\ref{sec:metric_improvement} we verify that even without any rejection, $\mu$-ScaleFace method improves the key metrics.

\subsection{Experimental setup}
\noindent\textbf{Datasets.} 
  For training all the models we use MS1MV2 dataset~\cite{deng2019arcface} which is the revised version of MS-Celeb-1M dataset~\cite{guo2016ms}. It contains the data about 85K identities with each identity having about 100 facial images. 

  For evaluation we use common IJB-C dataset~\cite{maze2018iarpa} (3.5K identities and 148.8K images) and cross-pose LFW~\cite{CPLFWTech} (2.3K identities, 6K images).

  We use LFW dataset~\cite{huang2020labeled} only for basic experiment, as all the considered models achieve almost perfect results for it.
  For preprocessing we follow pipeline similar to~\cite{meng2021magface}.


\noindent\textbf{Uncertainty estimation approaches.}
  We consider the following uncertainty estimation approaches ranging from the baseline norm of ArcFace embedding to the most recent and advanced methods:
  \begin{itemize}
    \item Norm~\cite{yu2020out}: the norm of the ArcFace embedding before normalization;
    
    \item PFE~\cite{Shi2019ProbabilisticFE}: Probabilistic Face Embeddings;
    
    \item MagFace~\cite{meng2021magface}: margin-based uncertainty estimate;
    
    \item ScaleFace (ours): an approach introduced in Section~\ref{sec:scale_theory}.

    \item $\mu$-ScaleFace (ours): an approach with the threshold $\mu$ selected using validation data, see Section~\ref{sec:modified_similarity}.
  \end{itemize}

  All the methods except for MagFace share the same trained ArcFace iResNet-50 backbone. MagFace~\cite{meng2021magface} trains backbone and uncertainty estimation module simultaneously, so it has different recognition quality compared to ArcFace and other methods for the same backbone architecture.

  For all the methods, we use hyperparameters and architectures suggested by their authors. For ScaleFace, we consider various design choices and their effects. See details in Supplementary Material. In particular, we take multilayer perceptron $s(\xv)$ with two hidden layers for the computation of scale values based on the embeddings from the penultimate level of the backbone.

\subsection{Qualitative Experiments}
\label{sec:qualitative_experiments}
\subsubsection{Face quality assessment}
  On the first step, we want to show that the proposed ScaleFace method provides perceptually reasonable uncertainty estimates. We divide images in the test sample by deciles of scale-based uncertainty $u(\xv)$ and uniformly randomly selected images from these deciles.
    
  Figure~\ref{fig:face_examples} presents five examples of images from the top decile and five examples of images from the bottom decile.
  As we see, the computed scale values provide an adequate representation of the quality of the image and consequently uncertainty for these images.
  Images with high uncertainty are blurry, dark, or only partly reveal the face.
  In contrast, images with low uncertainty allow easy identification of a depicted person.

\subsubsection{Comparison of uncertainties distributions for different datasets}
  Figure~\ref{fig:uncertainty_hist_open_set} presents comparison of histograms of confidence values for different considered datasets.
  For better presentation we apply Box-Cox transformation with $\lambda = 3$ to predicted scale values and linearly normalize results in a similar way for all datasets to constrain produced confidence values to the interval $[0, 1]$.
    
  For LFW and MS1MV2, the histograms are pretty close to each other with the apparent shift towards more confident decisions. For MS1MV2 it is due to the fact that this dataset was used for training, while LFW dataset is known to be relatively easy one for face recognition. At the contrary, for IJB-C the histogram is shifted to the left. It means that we are less certain about decisions for images in this dataset as it is really complicated real-world dataset. The presented results correlate with community~agreement~\cite{Shi2019ProbabilisticFE,meng2021magface} about complexity for these datasets.

\subsection{Reject verification}
\label{sec:reject_verification}
\subsubsection{Reject verification metric}
\label{sec:open_set_metrics}
  We consider so-called \textit{reject verification} evaluation procedure as a main tool to assess the quality of uncertainty estimates in the context of open-set recognition. We describe it in details below.

  We consider a test dataset $D_{test} = \{(\xv_{i1}, \xv_{i2}), y_i\}_{i = 1}^{\sS}$ consisting of pairs of images $(\xv_{i1}, \xv_{i2})$ and labels indicating whether these images belong to one identity ($y_i = 1$) or not ($y_i = 0$). 
  For each sample from the dataset the backbone assigns a similarity score $p_i = \langle\ev(\xv_{i1}), \ev(\xv_{i2})\rangle$. 
  Thus, we have predictions $p_i$ and target labels $y_i$ and can compare them via metrics for the binary classification problem. 
  In this work, we use the true acceptance rate for a fixed false acceptance rate TAR@FAR that is a common metric for the face verification task.

  For each image $\xv$, uncertainty estimator assigns a value $u(\xv)$ that represents the uncertainty of the backbone in the predicted embedding. 
  For a pair of images $\xv_1, \xv_2$ we get the uncertainty $u(\xv_1, \xv_2)$ as geometric mean of the uncertainties $u(\xv_1)$, $u(\xv_2)$.

  We expect that pairs of images with high uncertainty have bigger chances to be verified incorrectly. 
  We filter out a fixed share $r \in [0, 0.5]$ of image pairs with the biggest value of uncertainty to get the subset $D_{test}^r$ and calculate the TAR@FAR metric on the remaining ones. 
  Then we plot the dependence of TAR@FAR on the share of rejected pairs $r$ and calculate the area under this curve.
  Better uncertainty estimates lead to a faster growth of the curve, as we reject more ``bad'' pairs and, thus, the area under curve is also bigger.

  When $r = 0$, we use the whole sample $D_{test}$ to get the metric, so the starting point of the curve is the same for all uncertainty estimates, if the backbone is the same.
  Ideally, for as honest as possible comparison of different uncertainty estimation approaches, we need to use the same backbone network to calculate embeddings and similarities.

  For models with different base accuracy the final AUC score can be misleading, as both model accuracy and quality of uncertainty estimation influence the final result. To address the problem, we use the same ArcFace backbone and cosine similarity in part of experiments (Figure~\ref{fig:scale_search_activation_2} and Table~\ref{tab:uncertainty_improvement}). For the results with modified similarity metric, see Section~\ref{sec:metric_improvement}.

\subsubsection{IJB-C reject verification}
  We perform reject verification evaluation procedure described in Section~\ref{sec:open_set_metrics}. Figure~\ref{fig:scale_search_activation_2} presents the comparison of the proposed ScaleFace method with the baselines that share the same ArcFace backbone. We note that in this experiment all the approaches use the same cosine distance to compute the similarity.

  We see, that scale-based uncertainty estimate is better for typical values of FAR: $0.0001, 0.001, 0.05$, as the corresponding rejection curve is higher than that for other approaches. 
  Besides 1-to-1 image verification, we use N-to-1 test protocol as well, where one embedding is a mean of a few image embeddings. On both setups ScaleFace-based uncertainties performs very well, see Table~\ref{tab:uncertainty_improvement}. Interestingly, the approach based on the norm of the embeddings is competitive for small rejection rates but its quality rapidly deteriorates for the larger ones, while PFE performs much better for a N-to-1 template setup.

  \begin{table}[t!]
    \centering
    \begin{tabular}{lcccc}
      \hline
      FAR & $0.0001$ & $0.001$ & $0.01$ & $0.05$ \\
      \hline
      Verification & \multicolumn{4}{l}{ArcFace backbone} \\
      \hline
      Random	& 0.8080 & 0.8640 & 0.9074 & 0.9346 \\
      Norm & 0.9064 & 0.9378 & $\underline{0.9608}$ & $\underline{0.9738}$ \\
      PFE & $\underline{0.9200}$ & $\underline{0.9418}$ & 0.9588 & 0.9698 \\ 
      MagFace & 0.9068 & 0.9318 & 0.9520 & 0.9652 \\
      ScaleFace (ours) & $\mathbf{0.9336}$ & $\mathbf{0.9554}$ & $\mathbf{0.9706}$ & $\mathbf{0.9794}$ \\

      \hline
      Template verific. & \multicolumn{4}{l}{ArcFace backbone}  \\
      \hline
      Norm & 0.8872 & 0.9206 & 0.9472 & $\underline{0.9642}$ \\
      MagFace &	0.8934 &	0.9192 & 0.9422 &	0.9586 \\
      PFE  & $\underline{0.9102}$ & $\underline{0.9332}$ & \underline{0.9524} & $\underline{0.9642}$ \\ 
      ScaleFace (ours) & \textbf{0.9166} & \textbf{0.9404} & \textbf{0.9578} & \textbf{0.9688} \\
    \hline
    \end{tabular}
    \caption{
    AUC under rejection TAR@FAR curve on IJB-C
    test dataset. The first part is for 1-to-1 image pairs. The second
    part is for template N-to-1 face verification. We run experiments for different FARs for rejection portions from $0$ to
    $0.5$. Best value in \textbf{bold} and second best value \underline{underscored}. Results are normed by optimal value.}
    \label{tab:uncertainty_improvement}
  \end{table}

  \begin{table}[t!]
    \centering
    \begin{tabular}{lcccc}
        \hline
        FAR &  0.005 & 0.01 & 0.1 \\
        \hline
        Random & 0.7850 & 0.8294 & 0.8862 \\
        Norm & 0.8855 & 0.9215 & 0.9534 \\
        PFE & \underline{0.8967} & \underline{0.9261} & \underline{0.9539} \\
        ScaleFace (ours) & \textbf{0.9106} & \textbf{0.9361} & \textbf{0.9613} \\
        \hline
    \end{tabular}
    \caption{AUC under rejection curve for different TAR@FAR values for rejection portions from $0$ to $0.5$ on Cross-pose LFW dataset. Best value is in \textbf{bold} and second best value is \underline{underscored}. Results are normed by optimal value.}
    \label{tab:auc_lfw}
  \end{table}

\subsubsection{Cross-pose LFW reject verification}
  One commonly used dataset for an open-set task is ``Labeled faces in the wild'' (LFW; \cite{huang2020labeled}). It is rarely used now as all the considered models achieve almost perfect results for it. There is harder variations of it, i.e. cross-pose and cross-age LFW~\cite{CPLFWTech}. In these datasets positive pairs has two semantically different photos, e.g. a person is much younger on one of the images or photos were made from completely different positions. We decided to run a test for reject-verification, as it would test how ScaleFace and other methods work with semantically hard photos.

  The results are shown in Table~\ref{tab:auc_lfw}. Even basic embeddings norm estimation for an uncertainty allow to get meaningful improvement and ScaleFace got the best results for all false acceptance rates.
  For easier interpretation we also provide table for TAR@FAR values with 20\% samples rejected in Table~\ref{tab:tar_improvement}. As we can see, by getting rid of relatively small number of samples, we could dramatically reduce the error. In real-world scenario we could ask to redo the photo or send questionable pairs to a human expert.

  \begin{table}[t!]
    \centering
    \begin{tabular}{lcccc}
        \hline
        FAR &  0.005 & 0.01 & 0.1 \\
        \hline
        Base value & 0.781 & 0.825 & 0.884 \\
        Norm & 0.879 & 0.926 & 0.959 \\
        PFE & \underline{0.884} & \underline{0.926} & \underline{0.959} \\
        ScaleFace (ours) & \textbf{0.907} & \textbf{0.945} & \textbf{0.970} \\
        \hline
    \end{tabular}
    \caption{TAR@FAR values with 20\% of samples rejected based on uncertainty for Cross-pose LFW dataset. The base value refers to a starting model TAR without rejection.}
    \label{tab:tar_improvement}
  \end{table}

\subsection{Similarity metric improvement on IJB-C}
\label{sec:metric_improvement}
  In this experiment, we explored the applicability of the ideas described in Section~\ref{sec:modified_similarity}. Here we compare simple cosine similarity $\langle \ev(\xv_1), \ev(\xv_2) \rangle$ and similarity score improved by uncertainty prediction: $\sqrt{s(\xv_1) s(\xv_2)} \bigl(\langle \ev(\xv_1), \ev(\xv_2) \rangle - \mu\bigr)$. 
  Table~\ref{tab:backbone_improvement} shows that $\mu$-ScaleFace provides an improvement of the backbone even without rejection. On the full test dataset it uniformly improves over ArcFace and outperforms PFE for larger values of FAR.

  \begin{table}[t!]
    \begin{tabular}{lcccc}
        \hline
        FAR & 0.0001 & 0.001 & 0.01 & 0.05 \\
        \hline
        ArcFace & 0.8043 & 0.8704 & 0.9116 & 0.9382 \\
        PFE & \textbf{0.8224} & \textbf{0.8759} & \underline{0.9181} & \underline{0.9485} \\
        MagFace & 0.7741 & 0.8543 & 0.9074 & 0.9423 \\
        ScaleFace (ours) & 0.8070 & 0.8703 & 0.9116 & 0.9379 \\
        $\mu$-ScaleFace (ours) & \underline{0.8157} & \underline{0.8713} & \textbf{0.9194} & \textbf{0.9558} \\
        \hline
    \end{tabular}
    \caption{TAR values for different FARs on IJB-C dataset. The results for end-to-end trained models with similarity metrics corresponding to each method.}
    \label{tab:backbone_improvement}
  \end{table}

\section{Text-to-Image Retrieval}
\label{sec:retrieval}

\begin{table}[t!]
    \begin{tabular}{l|cc|cc}
        \hline
        Dataset &
        \multicolumn{2}{c|}{Conceptual Captions} & \multicolumn{2}{c}{COCO}\\
        \hline
        Metric & AUC & AUC@1 & AUC & AUC@1\\
        \hline
        CLIP & 0.1590 & \underline{0.1759} & 0.1132 & \underline{0.1186} \\
        Norm & \underline{0.1616} & 0.1745 & \underline{0.1143} & 0.1184 \\
        PFE & 0.1592 & 0.1747 & 0.1133 & 0.1183\\
        $\mu$-Scale & \textbf{0.2033} & \textbf{0.1926} & \textbf{0.1874} & \textbf{0.1611}\\
        \hline
    \end{tabular}
    \caption{Comparison of pretrained CLIP with its uncertainty-based modifications in text-to-image retrieval task.}
    \label{tab:rettieval_results}
  \end{table}
    
\subsection{Method}
  The developed method of uncertainty estimation can be applied to any metric learning-based task. To show that, we utilize ScaleFace for text-to-image retrieval.

  Here we need to retrieve images from a gallery relevant to a text query. A typical solution is to train text and image encoders that map inputs to the same embedding space. Then we calculate similarity distances for each caption-image pair and retrieve only close neighbors. The target metrics in this task are precision and recall. We need to retrieve as many relevant images as possible while keeping the number of incorrectly retrieved images low.

\noindent\textbf{Modifying similarity with uncertainty.} 
  We want to modify the original cosine similarity metric using uncertainty values computed on the embeddings. For that, two separate uncertainty-predicting heads $s_1, s_2$ are trained, one for text $\xv$ and one for image $\yv$ embeddings. Then we use predicted uncertainty values $s_1 = s_1\bigl(\ev(\xv)\bigr), s_2 = s_2\bigl(\ev(\yv)\bigr)$ to modify the original distance like this
  \begin{equation}
    \sqrt{s_1 \cdot s_2}\bigl(\langle \ev(\xv), \ev(\yv) \rangle - \mu\bigr),
  \label{eq:modified_cos}
  \end{equation}
  where $\mu$ is a threshold that separates negative and positive pairs computed similar to the one in the recognition task, see Section~\ref{sec:modified_similarity}. Our experiments show that the value of $\mu$ can be calculated on the training data. Thus, no additional validation data are required.

\subsection{Experiments}
\subsubsection{Experimental setup}
\noindent\textbf{Datasets.}
  There are two datasets used in our experiments: Conceptual Captions~\cite{sharma-etal-2018-conceptual} and COCO~\cite{10.1007/978-3-319-10602-1_48}. Both provide image-caption pairs and are quite popular in retrieval benchmarks. We train and test our models on the same dataset.
\noindent\textbf{Models.}
  We take a pretrained CLIP ViT-B-32~\cite{radford2021learning} as an encoder backbone and train MLP uncertainty predicting heads over its embeddings. Our experiments with heads' architectures revealed a very small impact on the final result. We take a multilayer perceptron with four hidden layers as an uncertainty predicting head and use the same architecture for both text and image heads.

\noindent\textbf{Uncertainty estimation approaches.}
  Some of the methods that we used in recognition can be utilized in retrieval. We test the following methods in our benchmarks:
  \begin{itemize}
    \item Norm~\cite{yu2020out}: modification of cosine similarity metric with norms of the embeddings before normalization like in~\eqref{eq:modified_cos};
    
    \item PFE~\cite{Shi2019ProbabilisticFE}: A direct implementation is possible like it was done in~\cite{karpukhin2022probabilistic} but with two separate variance predicting heads both for text and image embeddings;

    \item $\mu$--Scale (ours): an extension of $\mu$-ScaleFace described above.
  \end{itemize}
\noindent\textbf{Evaluation protocol.}
  First, we test all methods in typical retrieval when all sufficiently close to the query objects are retrieved. Setting different decision thresholds we get an approximation of the precision-recall curve. One can consider the area under this curve (Pr-Re AUC) as the target metric.

  Second, we test all the methods in limited retrieval setting when we allow to retrieve at most one object per query. This setting seems reasonable since our datasets provide only one image per caption. Here too we can plot precision-recall curves and compute the areas under them as quality metrics (Pr-Re AUC@1).

\subsubsection{Experimental results}
  We evaluate all the methods with Pr-Re AUC and AUC@1 metrics. Our experiments demonstrate superiority of the $\mu$-Scale algorithm over the competitors (see Table~\ref{tab:rettieval_results}) as Norm and PFE methods fail to improve the baseline. It seems that norms of CLIP's embeddings are not good measures of uncertainty and PFE cannot learn anything from pairs of embeddings.


\section{Related Work}
\label{sec:open_set_sota}

\noindent\textbf{Open-set and face recognition.}
  Open-set recognition quality significantly increased after the introduction of advanced loss functions in recent years. The survey~\cite{zheng2016person} documents how the field changed since the introduction of deep learning models, while a more recent survey~\cite{wang2021deep} demonstrates that more accurate treatment of embedding space can give even more impressive results with ArcFace~\cite{deng2019arcface} and subsequent works.
  A gentle introduction to the topic is given~\cite{hav4ik2021deepmetriclearning}. 
  We will focus on more recent works showing that ArcFace loss function and cosine distance in embeddings space lead to superior results. This fact is supported by the results of recent open-set competitions~\cite{weyand2020google} and~\cite{productMatching2020}.
    
\noindent\textbf{Uncertainty estimation for open-set recognition.}
  However, there are only a few works related to uncertainty estimation for open-set and face recognition, see recent surveys~\cite{abdar2021review,fu2022deep}. The two general ideas from uncertainty estimation for deep learning models can be seen in a variety of approaches: (i) use predicted maximum probability as a measure of confidence and (ii) train a separate head based on already obtained embeddings from a neural network backbone that would predict the uncertainty directly. 
    
  Probably the most well-known and important work so far is the one on Probabilistic Face Embeddings model~\cite{Shi2019ProbabilisticFE}. It demonstrates the high quality of uncertainty estimates and improves the quality of face recognition. It was also shown to improve models in general open-set recognition setup~\cite{karpukhin2022probabilistic} if compared to non-probabilistic approaches. The paper~\cite{Shi2019ProbabilisticFE} suggests to use an ArcFace backbone and train a separate network head to predict the variances of embeddings. Subsequently, it modifies the distance between embeddings using probabilistic model that takes these variances into account.

  Other papers also use ArcFace as a base approach and build on top of it.
  In particular, several recent papers on face recognition~\cite{ha2020google,yu2020out,meng2021magface} carry an idea of using not only the direction of the embedding vector as in ArcFace~\cite{deng2019arcface} but also its norm. 
  The paper~\cite{meng2021magface} considers margin term in the ArcFace loss function as the measure of confidence: large predicted margins correspond to high certainty of predictions.
  The paper~\cite{ha2020google} also proposes an approach to boost ArcFace model by selecting class-aware margins during training.
  Finally, intuition tells us that the more meaningful features are produced by a backbone, the more elements of the embedding vector will have high value. Even l2-norm of embedding from ArcFace~\cite{deng2019arcface} was shown to be a pretty strong baseline for this task in~\cite{yu2020out}.

\noindent\textbf{Scale selection in softmax.}
  Scale selection approach developed in this paper can be seen as setting a specific temperature for the loss function. 
  Selection of a single temperature for the whole dataset seems to be an important factor for a deep learning model calibration in general~\cite{guo2017calibration,minderer2021revisiting} and out-of-distribution (OOD) detection~\cite{liang2018enhancing}.

  The scale selection turns to be important for open-set recognition too~\cite{zhang2019adacos} with optimal value depending on the resulting embedding space and training epoch.
  Local temperature parameters specific for each image and part of an image have been used to improve calibration for the image segmentation problem~\cite{ding2021local}. 
  Note, that all these approaches target calibration quality improvement and require additional validation sample for temperature parameters estimation, thus complicating the overall model pipeline. 
  More recent works on OOD detection focus on the scale values prediction via a separate simple neural network~\cite{techapanurak2020hyperparameter,hsu2020generalized}.
  They highlight the importance of scale optimization for OOD detection. The work~\cite{techapanurak2020hyperparameter} suggests to learn scale (temperature) in image classification setup. However, they consider scale only on the training stage and do not use it on evaluation stage. The work~\cite{hsu2020generalized} makes a step further by using the learned scale values as a measure of uncertainty useful for OOD detection. However, they do not consider the open-set problem statement, where we achieve important benefits from considering the modified similarity metric between embeddings.
  
\noindent\textbf{Uncertainty estimation for retrieval.}
  The success of probabilistic approaches in face recognition inspired a number of similar approaches in retrieval. Usually these methods are tested in class-disjoint image-to-image retrieval benchmarks \cite{karpukhin2022probabilistic}. Among the most prominent approaches are HIB \cite{Oh2018ModelingUW}, PFE \cite{Shi_2019_ICCV}, DUL \cite{Chang_2020_CVPR}, SCF \cite{Li_2021_CVPR} and VMF-loss \cite{Scott_2021_ICCV}. As far as we know none of these methods has been implemented to text-to-image retrieval, especially on CLIP-sized models and large datasets.
    


\section{Conclusions}
\label{sec:open_set_conclusions}
  In this work, we introduce a new uncertainty estimation method, ScaleFace, designed for the open-set recognition and retrieval problems. The idea of the approach is to estimate the scale value in ArcFace~\cite{deng2019arcface} loss function for every input object and to use it as a confidence measure. Additionally, we introduce the modification to cosine distance based on the computed scale values.

  Our experiments examine the proposed measure and state-of-the-art uncertainty estimates from various angles and provide the detailed comparison of considered methods. Considered ScaleFace versions demonstrate significant improvement over PFE~\cite{Shi2019ProbabilisticFE} and other  baselines without any computationally-demanding modifications using instead a separate head of the neural network for scale estimation.  All the code to reproduce the experiments is available at \url{https://github.com/stat-ml/face-evaluation}. 
  
  Based on the conducted experiments, we recommend to use ScaleFace method for lightweight uncertainty estimate in open-set recognition and retrieval problems. It is fast to train, keeps inference time practically the same and provides superior uncertainty quality metrics.

  \noindent\textbf{Potential Negative Societal Impact.} 
  The developed algorithm can be used as part of a face recognition pipeline. Unfortunately, face recognition is often used for surveillance with potentially negative influence, especially in the case of more oppressive governments and organizations. We acknowledge it, but we believe there are more good applications for the technology: biometric security, information retrieval, and open-set entities search. We hope it will bring more good to the world.

  \noindent\textbf{Acknowledgments.} The research was supported by the Russian Science Foundation grant 20-71-10135.

{\small
\bibliographystyle{ieee_fullname}
\bibliography{openset}
}

\newpage 
\appendix
\section*{Supplementary material}
\label{sec:supplementary_material}
  \appendix 
\section{A general model for scale-based open-set recognition}
\label{sec:why_scale_face}

  There are many approaches aimed at uncertainty estimation for metric learning.
  In this section, we explain, why ScaleFace is pretty well suited for this task given the common assumptions about the metric space.

\begin{figure}
    \centering
    \includegraphics[width=0.35\textwidth]{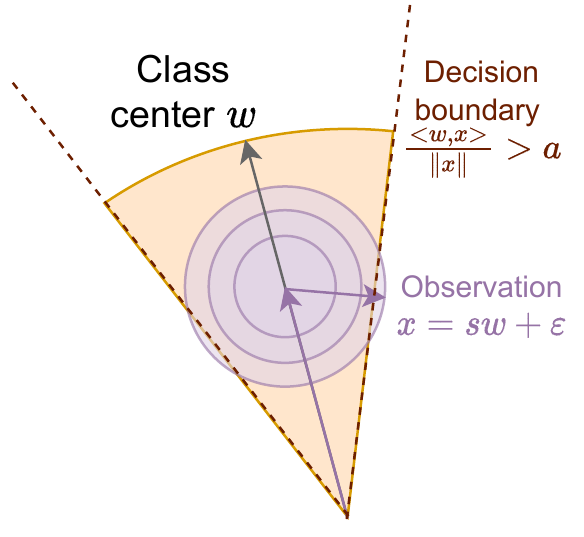}
    \caption{Data in embeddings space for a particular class. Lower scale values $s$ leads to higher error probabilities}
  \label{fig:scale_face_data_generation}
  \end{figure}

  Let us start with listing corresponding assumptions:
  \begin{itemize}
    \item[(A1)] For each class we have a single vector that represents the center of the class in the embeddings space. It is $\wv = \wv_i$ for $i$-th class. Due to used similarity measure, we assume that the vector has the unit $l_2$ norm: $\|\wv\|_2 = 1$.
    
    \item[(A2)] For large enough number of classes we can say, that a vector $\xv$ from embedding space belongs to $i$-th class, if $t = \wv^T \xv / \|\xv\|_2 > a = a_i$, where $a_i$ is a threshold for $i$-th class.
    
    \item[(A3)] Our observations in embedding space $\xv = s \wv + \epsv$, where $s$ is the scale and error $\epsv \sim \mathcal{N}(\mathbf{0}, \sigma^2 I)$.
  \end{itemize}
  The data generation scheme that follows from these assumptions is in Figure~\ref{fig:scale_face_data_generation}.

  Assumptions (A1)-(A2) are natural for multiclass ArcFace models based on cosine similarity. Assumption (A3) is typical for uncertainty estimation papers, where each object is associated with a multivariate Gaussian distribution in an embeddings space~\cite{Shi2019ProbabilisticFE}.

  Given these assumptions, we can derive, that the error probability for such a classifier is a function of the norm~$s = \| \xv \|_2$.
  We can approximate $t$ in the following way:
  \[
    t \approx \hat{t} = \sum_{i = 1}^d w_i (s w_i + \varepsilon_i) / s,
  \]
  as perturbation by $\epsv$ is small compared to the vector $\wv$.
  \[
    \sum_{i = 1}^d w_i (s w_i + \varepsilon_i) / s = 1 + \frac1s \sum_{i = 1}^d w_i \varepsilon_i. 
  \] 
  The noise values $\varepsilon_i$ are independent, the distribution of $t$ is close to the following Gaussian distribution:
  \[
    \hat{t} \sim \mathcal{N}\left(1, \frac{\sigma^2}{s^2}\right),
  \]
  as $\| \wv \|_2 = 1$.
  So, the variance of this distribution is proportional to~$\frac1{s^2}$.
  We are interesting in the error probability: 

  \begin{equation} \label{eq:error_probability_scale}
    P(t < a) \approx 2 (1 - \Phi(\frac{s}{\sigma} (1 - a))).
  \end{equation}

\begin{figure}
    \centering
    \includegraphics[width=0.4\textwidth]{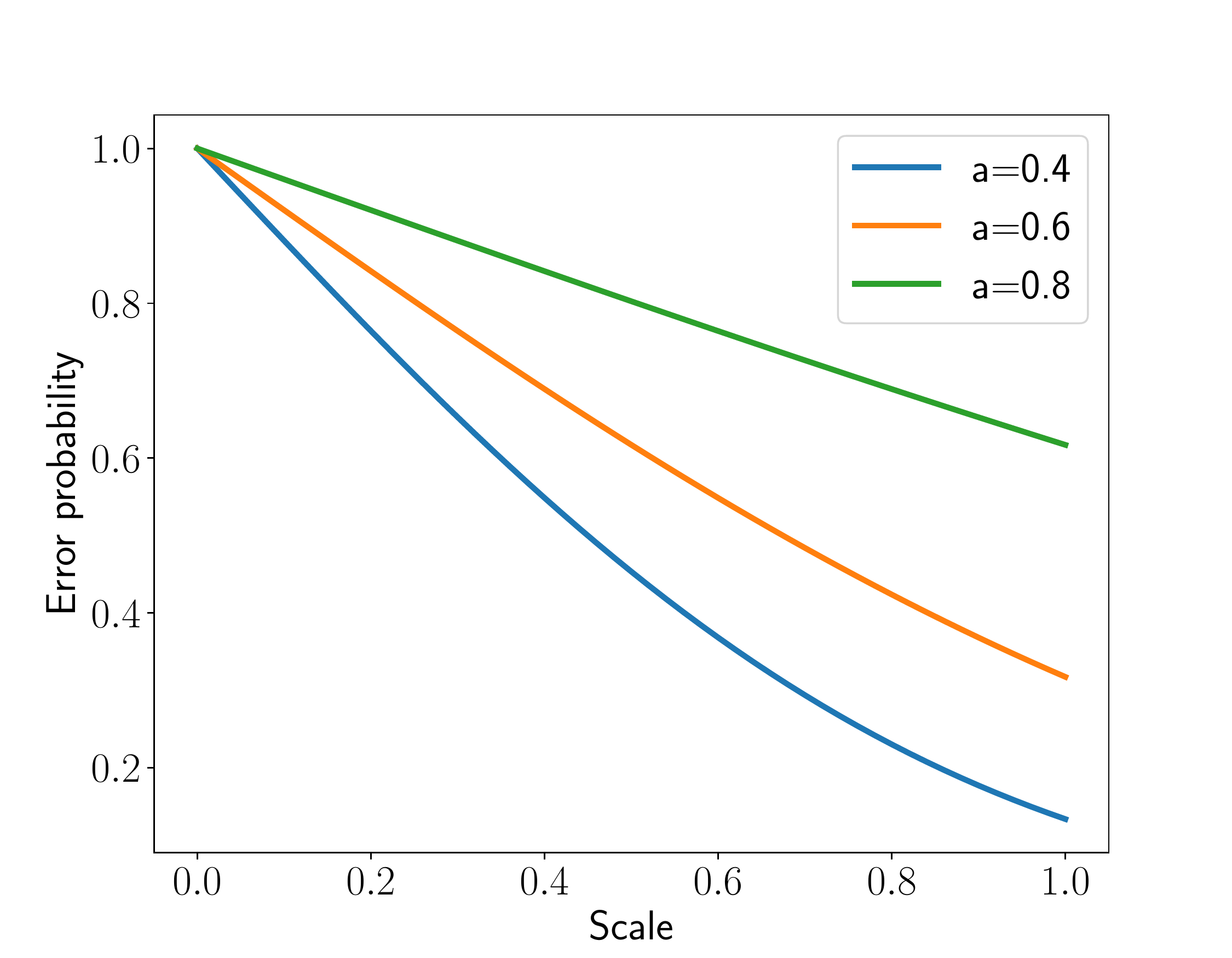}
    \caption{Error probability for different decision thresholds $a$ and different scales $s$ as predicted by equation \ref{eq:error_probability_scale}. We could see that bigger scale leads to lower error probability.}
  \label{fig:error_prob}
 \end{figure}

  It is easy to see, that the error probability is monotonic in $s$: as the scale $s$ increases, the error probability decreases. 
  An example of obtained error probabilities for a particular decision boundaries is in Figure~\ref{fig:error_prob}.
   
\section{$\mu$-ScaleFace algorithm training pipeline}
\label{sec:mu_scaleface}

  Below we present the pipeline that allows reasonable solution of such problem and leads to $\mu$-ScaleFace algorithm:
  \begin{enumerate}
    \item We assume that a validation dataset $D_{val} = \{(\xv_{i1}, \xv_{i2}), y_i\}_{i \in \mathrm{val}}$ is available. 
    Validation set should be different from the training one and will be used to find the class-separating threshold. 

    \item For each pair in the validation dataset, we calculate cosine similarity $d_i = \langle \ev(\xv_{i1}), \ev(\xv_{i2}) \rangle$. 
        Our validation sample consists of two subsamples: $D_{val}^+$ and $D_{val}^-$ containing positive and negative pairs correspondingly.
        Then we calculate mean similarity for $\mu^+ = \frac{1}{|D_{val}^+|}\sum_{i \in D_{val}: y_i = 1} d_i$ positive and $\mu^- = \frac{1}{|D_{val}^-|}\sum_{i \in D_{val}: y_i = 0} d_i$ negative class for validation data. Then we average these two class centres to get a class-separating threshold $\mu = \frac{1}{2} (\mu^+ + \mu^-)$.

    \item We use the computed value of $\mu$ to modify the similarity measure according to equation in the main text.
    For example, for a pair of objects \((\xv_{j1}, \xv_{j2})\) we can compute $\tilde{d}_j = s(\xv_{j1}, \xv_{j2}) (\langle \ev(\xv_{j1}), \ev(\xv_{j2}) \rangle - \mu)$ and use these new similarities $\tilde{d}_j$ 
    for open-set recognition.
  \end{enumerate}

\section{Detailed reject verification protocol}
  To sum up, we have the following steps for the comparison of uncertainty estimates based on the reject verification:
  \begin{enumerate}
    \item Select a backbone and a test sample of pairs $D_{test}$.

    \item Select an uncertainty estimate for a single image $u(\xv)$ and for a pair of images $u(\xv_1, \xv_2)$ based on that for a single image.

    \item For a grid $r_0 = 0 \le r_1 \le \ldots \le r_k$ get TAR@FAR rejection curve:
    \begin{enumerate}
      \item reject $r_i$-th share of objects with highest values of $u(\xv_1, \xv_2)$ to get $D_{u, test}^{r_i}$;

      \item get the quality metric TAR@FAR for $D_{u, test}^{r_i}$.
    \end{enumerate}

    \item Calculate area under TAR@FAR rejection curve: higher values correspond to better models.
  \end{enumerate}

\section{Additional Experiments}
  In this section we provide additional examination of ScaleFace approach and design choice studies for it.
  For all the experiments we use IJB-C as the test dataset.

\subsection{Mean faces for confidence bins}
  To get an additional sanity check of our approach, we provide mean faces for each uncertainty bin similar to~\cite{meng2021magface}.
  We predict confidence for each image from the IJB-C dataset. 
  Then we split images into 8 bins, according to the confidence and averaged images in each bin pixel-wise. 

\begin{figure*}[h!]
    \includegraphics[width=1.0\textwidth]{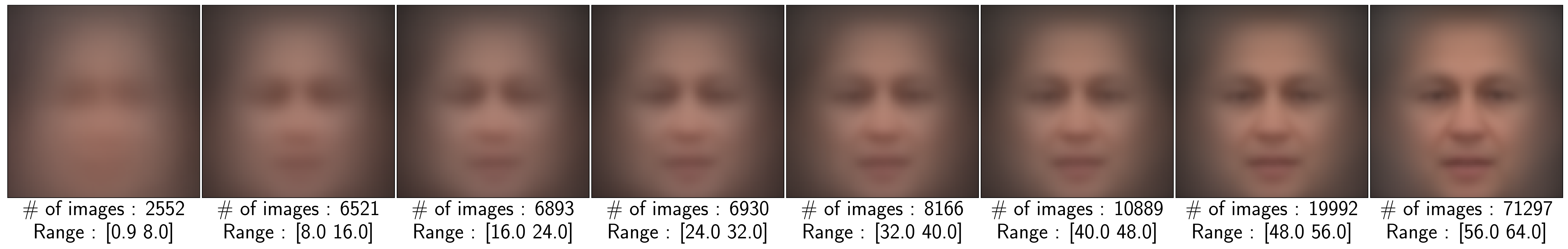}
    \centering
    \caption{Mean faces for each confidence bin for the IJB-C dataset. Confidence bean is a result of ScaleFace application to the dataset.}
    \label{fig:confidence_mean_face}
  \end{figure*}

  Figure~\ref{fig:confidence_mean_face} presents resulting mean images.
  We see, that the mean for the least confident images is blurry, while for the mean of the most confident images we see a readable face.
  We argue, that most certain images are typically mug shots with clearly distinguishable facial features, while among images with low confidence there are many images with profile view, blurry or corrupted in other ways.

\subsection{Design choice studies}
  No principled choice exists for the scale projection head. 
  We conduct an ablation study with different options for architecture and activation function.
  For this part of the experiments, the weights of the backbone are frozen with pre-trained ArcFace model weights. The architecture of the head consists of fully-connected layers with ReLU activations, which predicts a scalar $a_i$ for each object $\xv_i$. We apply the activation function to $a_i$ to ensure the positivity of the resulting scale $s_i$.

\begin{figure*}[b!]
    \includegraphics[width=0.95\textwidth]{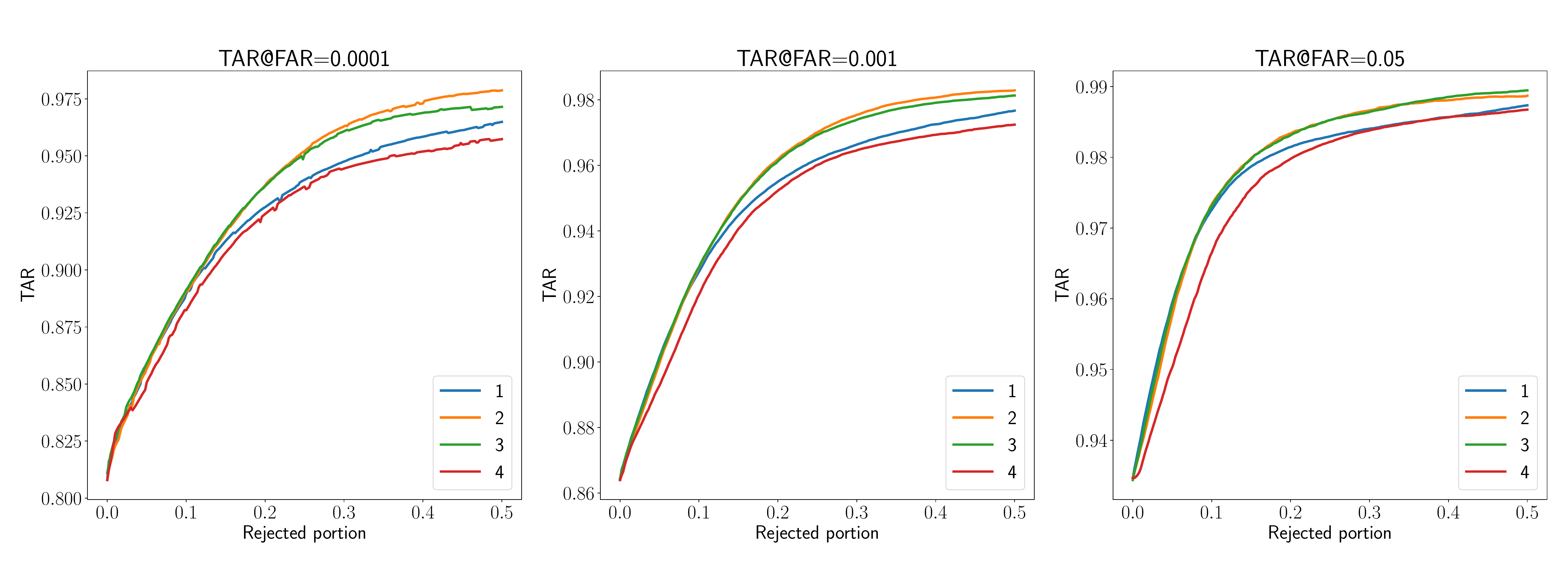}
    \centering
    \caption{Rejection curves for projection heads with different number of layers from $1$ to $4$. Each figure refers to different value of FAR in TAR@FAR metric.}
  \label{fig:scale_search_mlp_depth}
  \end{figure*}

\paragraph{Number of layers for scale prediction.}
  First, we compare different number of fully connected layers in our scale prediction head $1, 2, 3, 4$ with $s_i = 32 \,\sigm(a_i)$ activation. Results are shown in Figure~\ref{fig:scale_search_mlp_depth}. The head with two hidden layers seems to be the best option.
  One layer seems to be not expressive enough for the problem at hand, while four layers lead to overfitting given the number of parameters involved.

\begin{figure*}[b!]
    \includegraphics[width=0.95\textwidth]{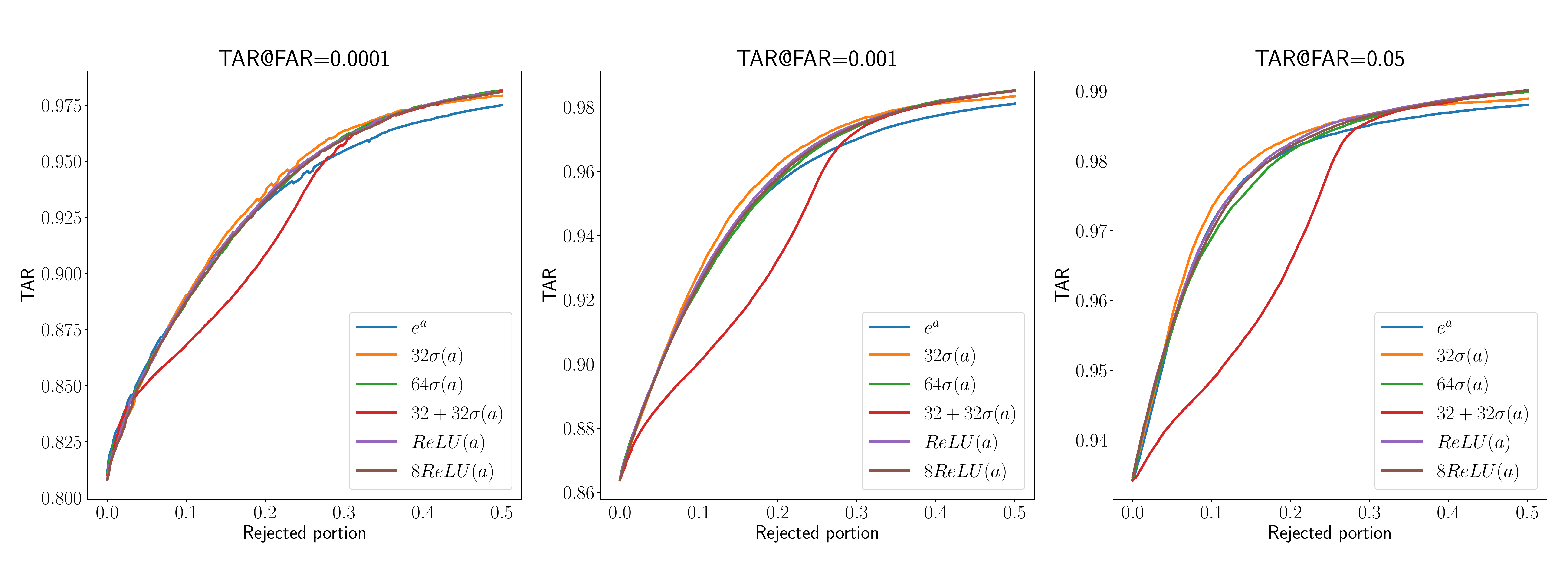}
    \centering
    \caption{In order to use MLP prediction as an uncertainty measure, we need to map prediction to $[0, +\infty)$. For this purpose we've sorted through several activation functions}
  \label{fig:scale_search_activation}
  \end{figure*}

\paragraph{Activation functions.} 
  Then, we go through several options for the activation function for the scale similar to used in the literature that produce non-negative values:
  \begin{itemize}
    \item $s_i = \exp(a_i)$;

    \item $s_i = c \, \sigm(a_i), c = 32, 64$;

    \item $s_i = 32 + 32 \, \sigm(a_i)$;

    \item $s_i = c \, \mathrm{ReLU}(a_i), c = 1, 8$.
  \end{itemize}
  The corresponding results are presented in Figure~\ref{fig:scale_search_activation}. 
  We see that the activation function $s_i = 64 \,\sigm(a_i)$ is the most stable option and provides a better quality, though performance differences are not very large compared to other choices. 
  
\begin{figure*}[h]
    \includegraphics[width=0.95\textwidth]{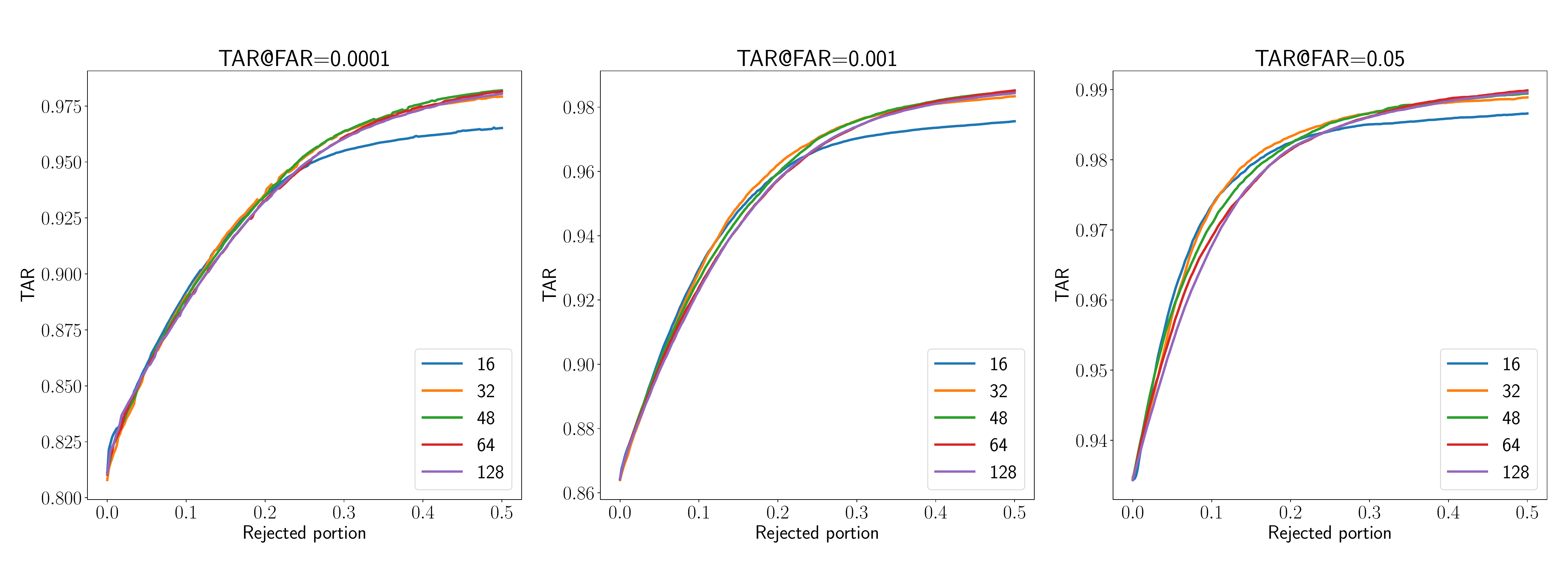}
    \centering
    \caption{Performance for different coefficient before $\mathrm{sigm}$ in ScaleFace scale head activation function.}
    \label{fig:scale_search_coefficient_s}
  \end{figure*}

\paragraph{Coefficient in the activation function.}
  To justify selection of the coefficient in front of the sigmoid activation function, we conducted an experiment to compare different possible options in Figure~\ref{fig:scale_search_coefficient_s}.
  We see, that all large enough values $\geq 32$ suit. 
  So, we selected $64$ as one that provides a stable performance.
  
  Thus, we ended up with the architecture with two fully-connected layers and $s_i = 64 \,\sigm(a_i)$ activation in all the experiments in the main paper if not specified otherwise.

\subsection{Template-based reject verification}
  
\begin{figure*}[t!]
    \includegraphics[width=0.9\textwidth]{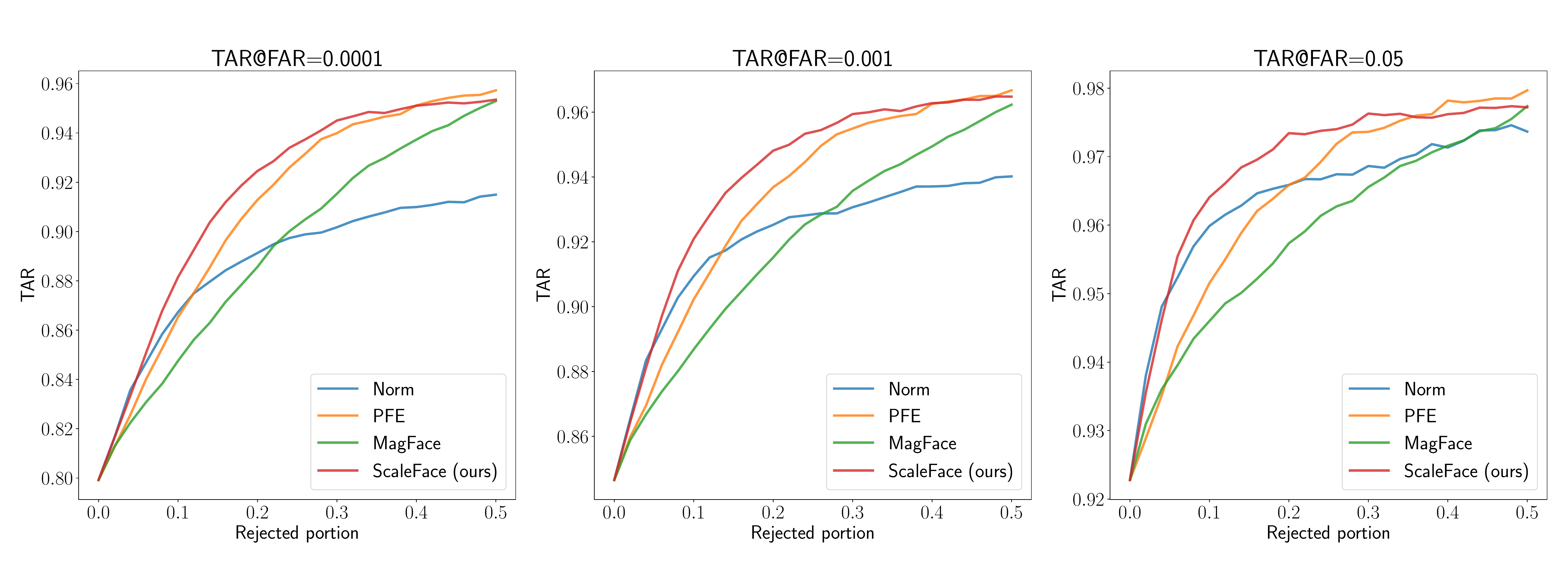}
    \centering
    \caption{Prediction with rejection for face templates verification. All methods use cosine distance.}
    \label{fig:face_templates_cosine}
  \end{figure*}

  We also conducted experiments based on templates approach to the open-set
  classification.
  The results are in Figure~\ref{fig:face_templates_cosine}.
  The improvement provided by our approaches is evident from these experiments for all considered TAR@FAR values.
  
\subsection{Modified distance experiments}
\begin{figure*}[t!]
    \includegraphics[width=\textwidth]{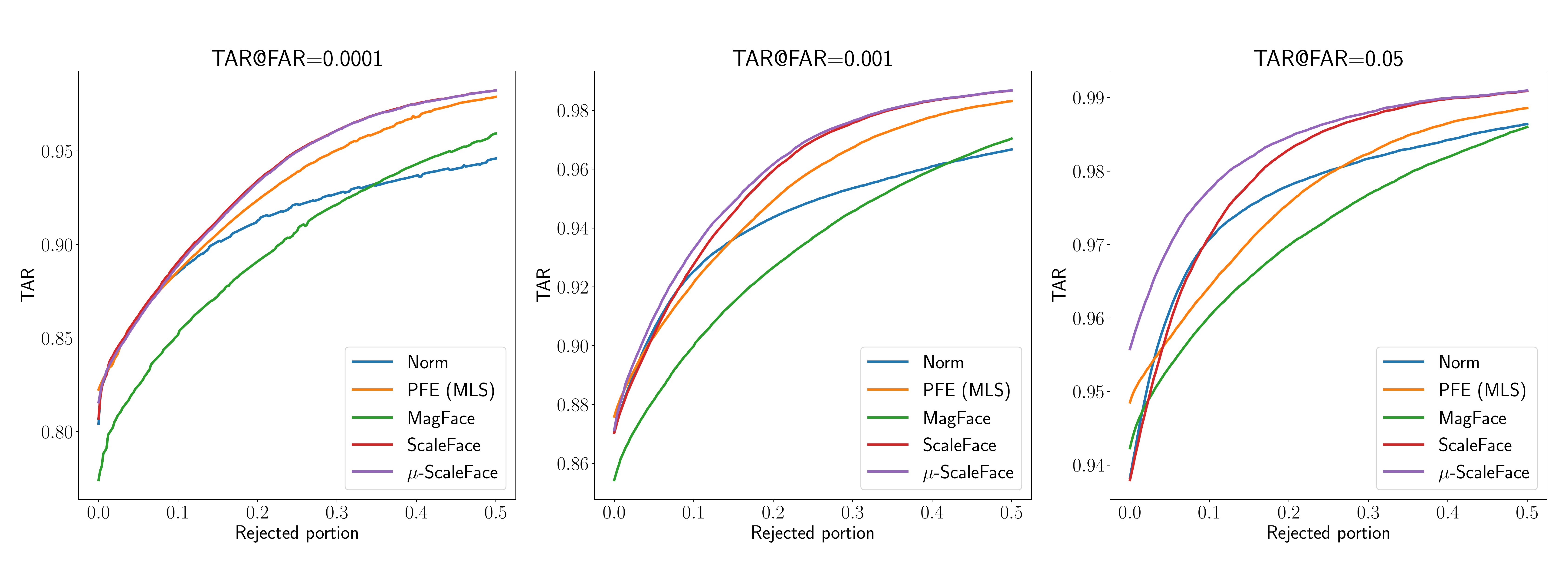}
    \centering
    \caption{Rejection curves for uncertainty estimation in verification task. PFE and $\mu$-ScaleFace tune the distance metric and MagFace tunes backbone, so the curves start at different points.}
  \label{fig:scale_verification}
  \end{figure*}

  Results of experiments with modified distance can be seen at Figure~\ref{fig:scale_verification}.

  \begin{table}[t!]
    \centering
    \begin{tabular}{lcccc}
      \hline
      TAR@FAR= & $0.0001$ & $0.001$ & $0.01$ & $0.05$ \\

      \hline
      Verification & \multicolumn{4}{l}{Trained backbones}  \\
      \hline 
      Norm & 0.9104 & 0.9418 & 0.9640 & 0.9764 \\
      PFE & 0.9274 & 0.9502 & 0.9654 & 0.9758 \\ 
      MagFace & 0.8972 & 0.93 & 0.9546 & 0.971 \\
      ScaleFace (ours) & $\mathbf{0.9366}$ & $\mathbf{0.9586}$ & $\mathbf{0.9732}$ & $\underline{0.9812}$ \\
      $\mu$-ScaleFace (ours) & $\underline{0.9324}$ & $\underline{0.9546}$ & $\underline{0.9724}$ & $\mathbf{0.983}$ \\

      \hline
      Template verific. & \multicolumn{4}{l}{Trained backbones}  \\
      \hline
      Norm & 0.8872 & 0.9206 & 0.9472 & 0.9642 \\
      MagFace & 0.8752 & 0.9092 & 0.9394 & 0.9594 \\
      PFE & \textbf{0.9192} & $\underline{0.9386}$ & \underline{0.9530} & $\underline{0.9644}$ \\
      ScaleFace (ours) & $\underline{0.9166}$ & \textbf{0.9404} & \textbf{0.9578} & \textbf{0.9688} \\
      \hline
    \end{tabular}
    \caption{AUC under rejection TAR@FAR curve for different TAR@FAR $0.0001$, $0.001$, $0.01$, $0.05$ for rejection portions from $0$ to $0.5$ with best value in \textbf{bold} and second best value \underline{underscored}. Results are normalized by optimal AUC value.}
    \label{tab:similarity_rejection_improvement}
  \end{table}

\subsection{$\mu$-ScaleFace on templates}
\begin{figure*}[t!]
    \includegraphics[width=\textwidth]{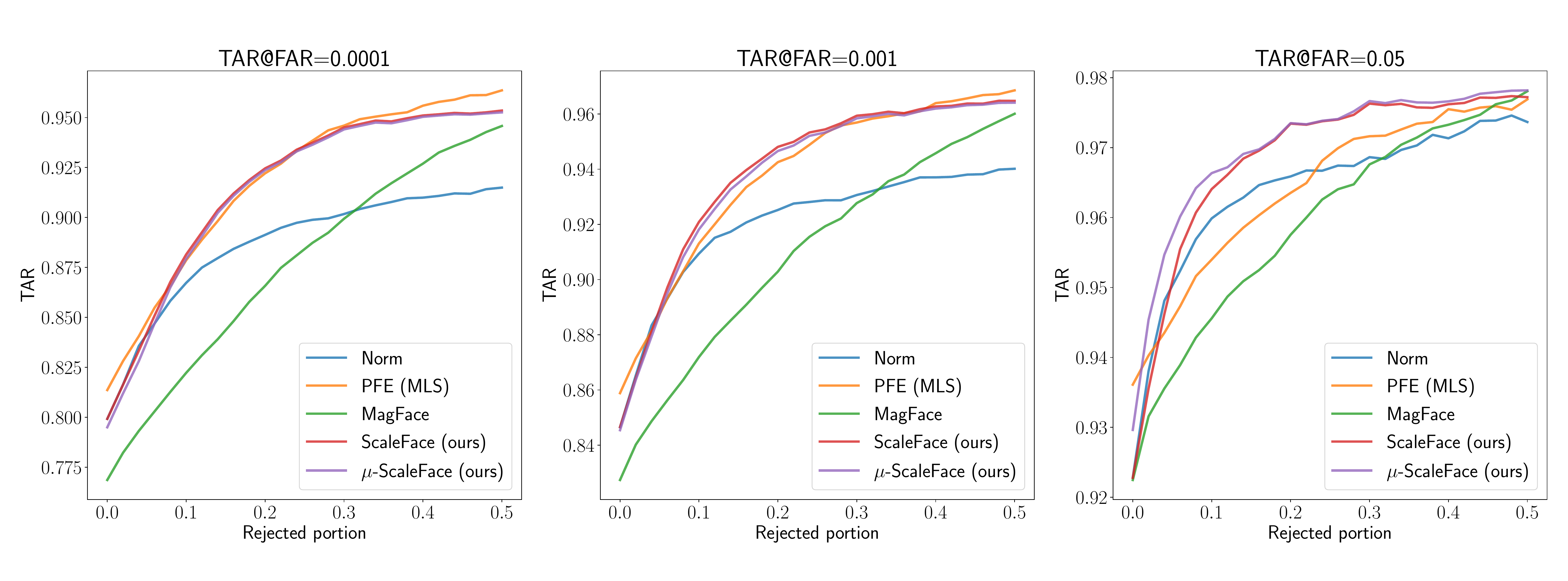}
    \centering
    \caption{Prediction with rejection for face templates verification. ScaleFace and MagFace use cosine distance, while PFE uses MLS. $\mu$-ScaleFace uses the improved distance. The improved distance allows to get better results for high values of FAR.}
    \label{fig:face_templates_mu}
  \end{figure*}

  We applied the improved ScaleFace metric for the experiments with templates described in the main text. 
  We used mean feature fusion with modified metric function and tuned $\mu$ to get $\mu$-ScaleFace. 
  The results are on Figure~\ref{fig:face_templates_mu}.
  The proposed natural modification allowed us to get some improvement, but only for high values of FAR.


\section{Experiments with text-to-image retrieval}

\begin{figure}[t!]
\includegraphics[width=0.9\linewidth]{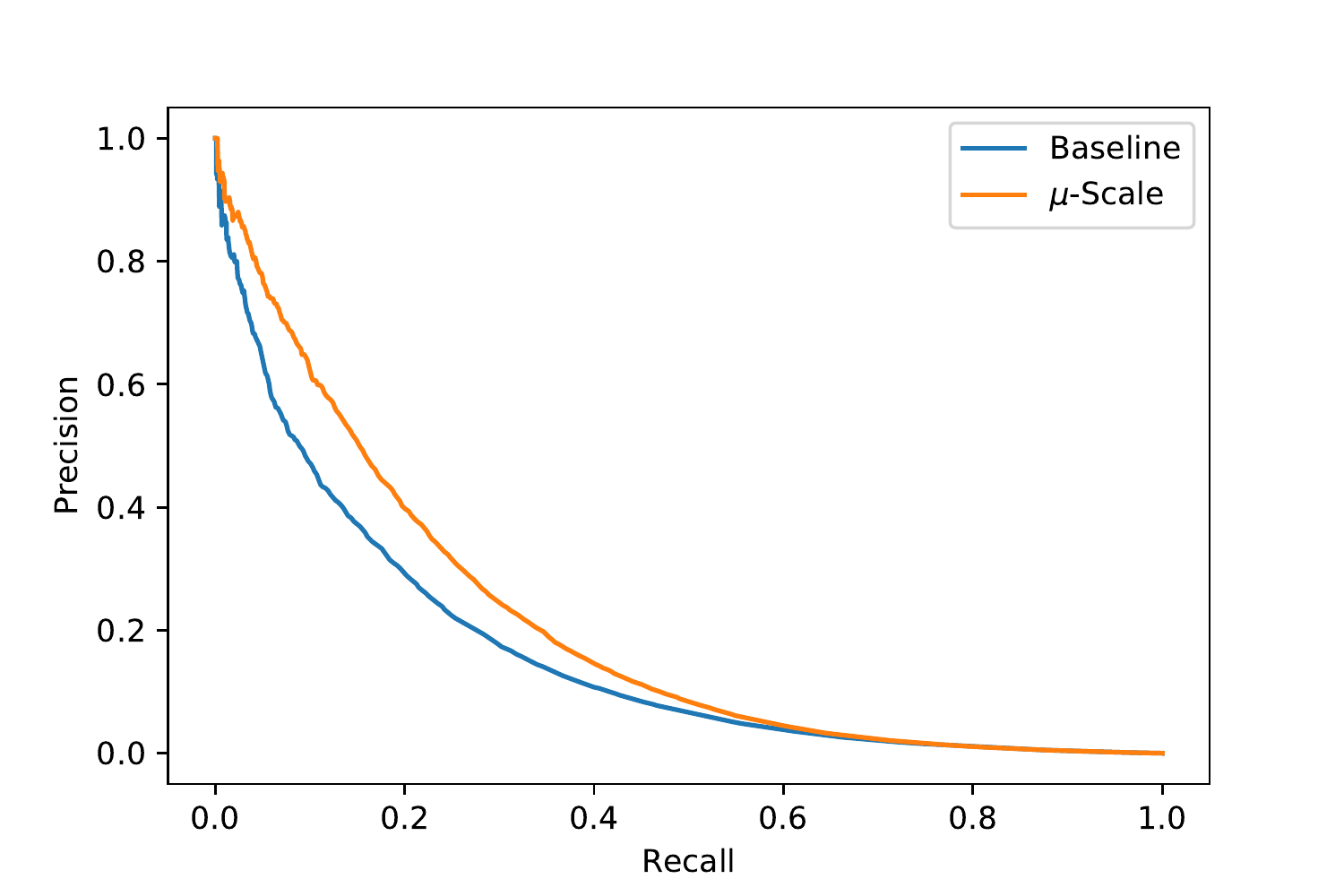}
\centering
\caption{Precision-recall curves for text-to-image retrieval on Conceptual Captions dataset.}
\label{fig:CC}
\end{figure}

\begin{figure}[t!]
\includegraphics[width=0.9\linewidth]{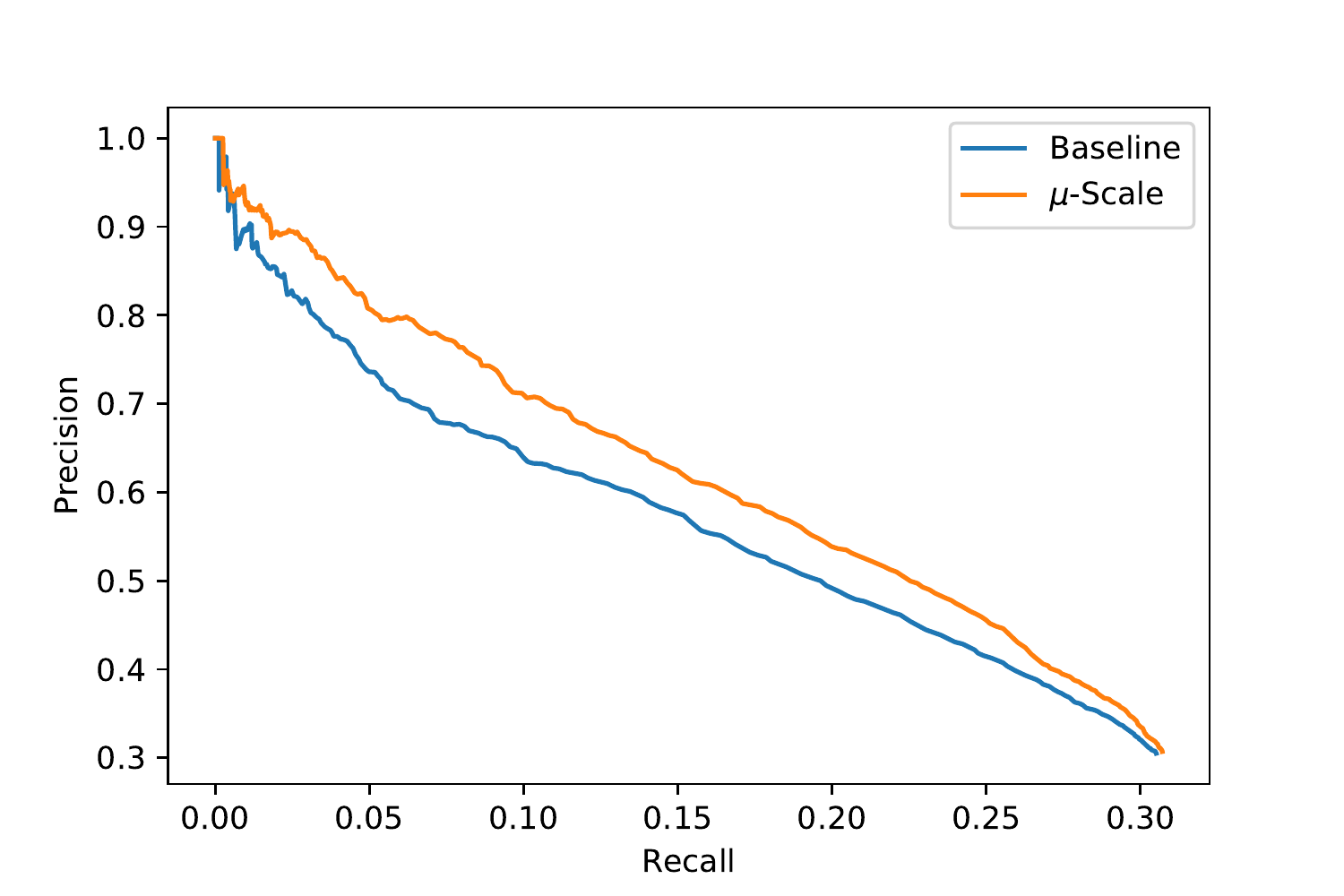}
\centering
\caption{Precision-recall curves for top1 text-to-image retrieval on Conceptual Captions dataset.}
\label{fig:CC_classification}
\end{figure}

\begin{figure}[t!]
\includegraphics[width=0.9\linewidth]{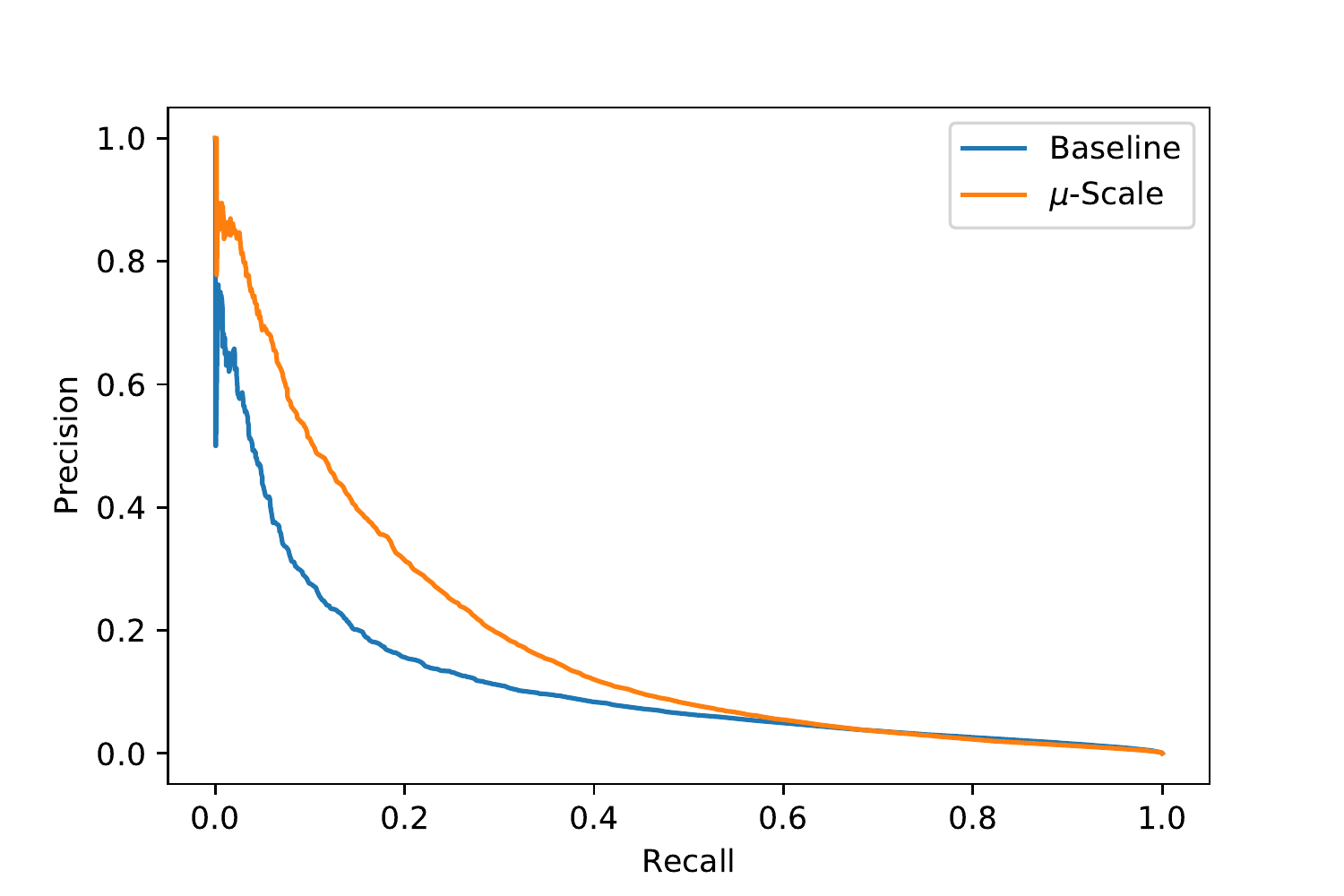}
\centering
\caption{Precision-recall curves for text-to-image retrieval on COCO dataset.}
\label{fig:COCO}
\end{figure}

\begin{figure}[t!]
\includegraphics[width=0.9\linewidth]{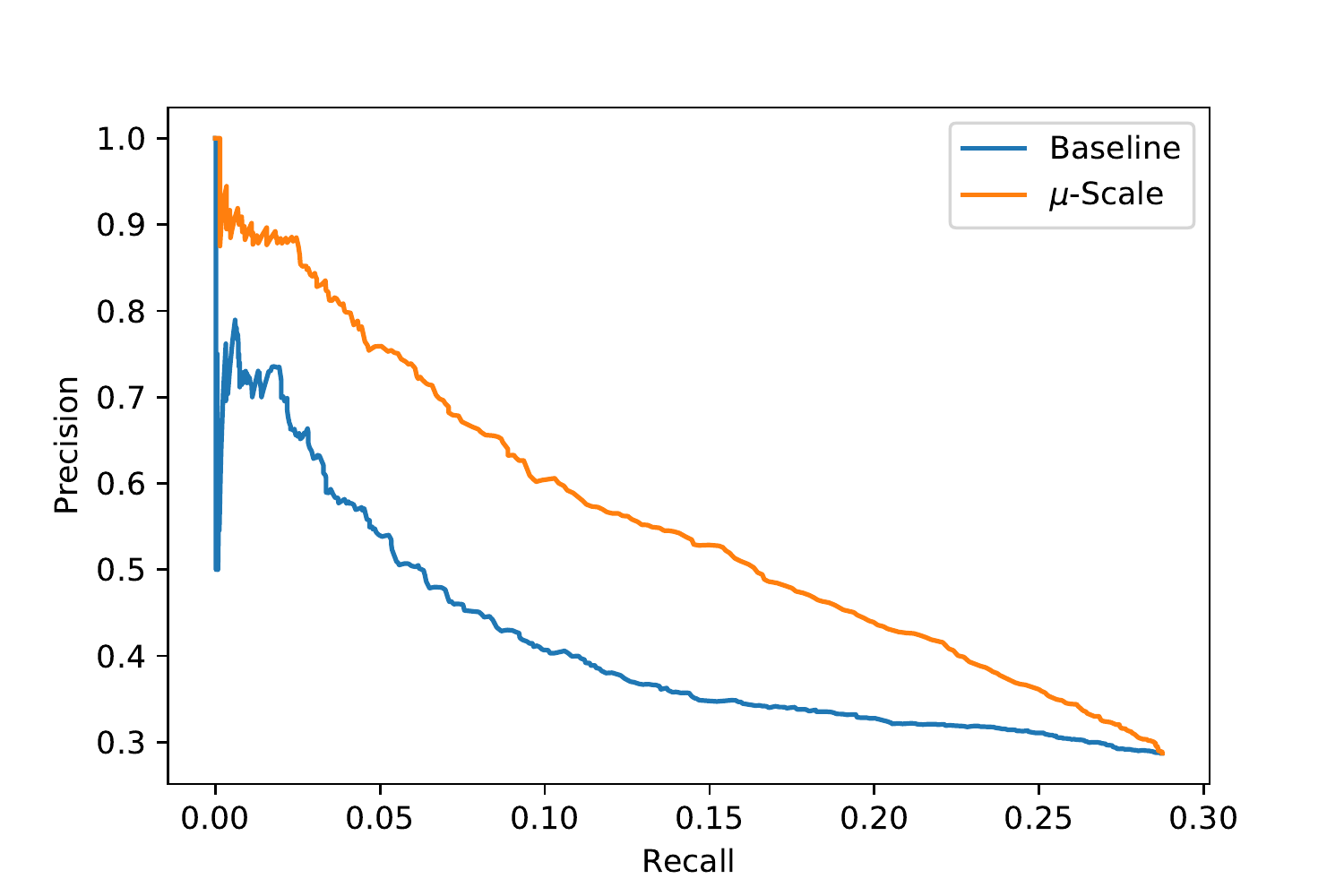}
\centering
\caption{Precision-recall curves for top1 text-to-image retrieval on COCO dataset.}
\label{fig:COCO_classification}
\end{figure}

Here we present precision-recall curves for CLIP baseline and our $\mu$-Scale solution (Figures \ref{fig:CC}, \ref{fig:CC_classification}, \ref{fig:COCO}, \ref{fig:COCO_classification}). The corresponding plots for Norm and PFE methods are omitted for clarity as they are very close to those for the baseline.
We see, that our approach provides a significant improvement over the baseline for all considered cases.

~

~




\end{document}